\documentclass[a4paper]{article}

\usepackage[pages=all, color=black, position={current page.south}, placement=bottom, scale=1, opacity=1, vshift=5mm, contents={DRAFT}]{background}

\usepackage[margin=1in]{geometry} 

\usepackage[sort&compress,numbers,square]{natbib}
\usepackage{subfigure}
\usepackage{booktabs}
\usepackage{multirow}
\usepackage{hyperref}

\title{TAI Scan Tool: A RAG-Based Tool With Minimalistic Input for Trustworthy AI Self-Assessment}
\author{Athanasios Davvetas$^1$ \and Xenia Ziouvelou$^1$\and Ypatia Dami$^1$ \and Alexios Kaponis$^1$ \and Konstantina Giouvanopoulou$^1$ \and Michael Papademas$^1$}

\date{
	$^1$National Centre for Scientific Research ``Demokritos'' \\ Institute of Informatics and Telecommunications \\ Athens \\ Greece \\ \texttt{\{tdavvetas, xeniaziouvelou, ypatia, akaponis, kgiouvano, papademasmichael\}@iit.demokritos.gr}\\
}

\begin{document}
\maketitle

\begin{abstract}
    This paper introduces the TAI Scan Tool, a RAG-based TAI self-assessment tool with minimalistic input. The current version of the tool supports the legal TAI assessment, with a particular emphasis on facilitating compliance with the AI Act. It involves a two-step approach with a pre-screening and an assessment phase. The assessment output of the system includes insight regarding the risk-level of the AI system according to the AI Act, while at the same time retrieving relevant articles to aid with compliance and notify on their obligations. Our qualitative evaluation using use-case scenarios yields promising results, correctly predicting risk levels while retrieving relevant articles across three distinct semantic groups. Furthermore, interpretation of results shows that the tool's reasoning relies on comparison with the setting of high-risk systems, a behaviour attributed to their deployment requiring careful consideration, and therefore frequently presented within the AI Act.    
\end{abstract}

\section{Introduction}
The rapid evolution of Artificial Intelligence (AI) has enabled its launch and use on numerous applications across the world offering economic and social benefits. However, its emergence has been associated both with hype and scepticism, regarding its implications for humans and society \cite{Ziouvelou2024}. Recently, the use and adoption of General Purpose AI (GPAI) systems has been increasingly growing, mainly due to the recent advancements in Large Language Models (LLMs), and their open accessibility. However, this wide adoption has uncovered a numerous new complex challenges related to ethical use and legal regulation of emerging AI systems \cite{Revolidis2024}. Therefore, we see that the technological evolution, amplifies the importance and need for Trustworthy AI (TAI), which is becoming increasingly applicable \cite{Li2023}.

The study of trustworthiness of AI systems, aims to cultivate trust in such systems. The High Level Expert Group (HLEG) of the European Commission \cite{EC2019} has provided a framework for Trustworthy AI (TAI), which consists of three components, namely, ethical, legal, and robust. Furthermore, it includes a set of TAI principles as requirements for the realisation of TAI. These components ensure that they safeguard human values and respect ethical principles (ethical), comply with all applicable laws and regulations (legal), and are robust from a technical and social perspective (robust). Furthermore, it states that these three components should be met throughout the entire system's lifecycle allowing our society to achieve ``responsible competitiveness''. Besides, adoption of the TAI principles and their evaluation with technical and non-technical means supports quicker, safer, and reliable adoption of responsible AI systems.

However, the process of incorporating and aligning with all three components, may take significant human and economic resources, especially for micro and small organisations and start-ups \cite{Bessen2022}. This can be attributed to both the need for specialised knowledge required for integrating and harmonising with these principles, as well as, the effort required for evaluation of AI systems across their lifecycle to ensure compliance. Small-Medium Enterprises (SMEs) or startups allocate the majority of their resources towards development of novel and competitive AI solutions in a highly competitive global market. In turn, although organisations strive for technological expertise, they tend to lack the legal or ethical know-how due to limitation in terms of resources or lack of in-house expertise \cite{Bessen2022}. Therefore, technical solutions that facilitate compliance with TAI principles, empower SMEs to achieve ``responsible competitiveness'' and bridge the gap of resources by reducing the resource burden through the process of self-assessing.

In this paper, we present the TAI Scan Tool, a tool based on the Retrieval Augmented Generation (RAG) framework for TAI self-assessment with minimalistic user-input. The current version of the system focuses on the legal aspects, aiming to ensure compliance with the EU Artificial Intelligence Act (AI Act)\footnote{Regulation EU 2024/1689, EU Artificial Intelligence Act, European Parliament, 2024}. We have provisioned the extension of the system to enable compliance with all TAI components, including legal, ethical, and robust. Existing products that enable compliance with the European regulatory landscape, often rely on rule-based systems to produce decisions and guidance to legal compliance. Domain or task specific rules are manually intensive, hard to maintain and prone to errors. In the constant changing landscape of regulatory frameworks, quick adaption to new amendments, guidelines, or new decisions is a critical feature for relevancy. Our system utilises a RAG system to overcome the challenge of error-prone and document specific rules. It also enables semantically rich reasoning and guidance towards compliance. The assessment process has two steps: (i) a pre-screening step that mostly aims to notify the user of non-acceptable practices and (ii) a legal TAI assessment according to the AI Act. It provides an initial assessment regarding the risk-level of the AI system according to the European AI Regulation. Furthermore, it provides relevant references that enable compliance with the corresponding obligations.

\section{Related Work}
In this section, we explore work related to ours, particularly on relevant approaches for LLM-based architectures in the legal domain and other legal AI assessment tools. Despite the recent advancements of LLMs, they still face the challenge of producing factually incorrect content, also known as ``hallucinating''. \citet{Huang2025} suggested a comprehensive taxonomy on hallucinations highlighting the key reasons for factually incorrect content. The authors suggest that hallucination are either a result of data, training, or inference. \citet{Gao2023} have showcased the use of the RAG framework as a way to deal with hallucination on LLMs in multiple domains and downstream tasks. The authors highlight especially that domain-specific or knowledge-intensive tasks require RAGs to overcome hallucinations. The process of TAI assessment is both a domain-specific as it involves content from domains such as, the law, ethics, technology, among others and a knowledge-intensive task. Therefore, the need to ground the reasoning and generation with knowledge from relevant documents is imperative. Additionally, we base our approach on RAG to provide an easily extensible but document-aware generation.

The use of the RAG framework in the legal domain has been accumulating attention to alleviate the challenges of plain LLMs in these applications. \citet{Kalra2024} introduce the HyPA-RAG, a hybrid and adaptive RAG configuration that includes dense, sparse and knowledge graph representations within the retrieval model. Additionally, it employs an adaptive adjustment of parameters that is based on the complexity of the query. \citet{Wiratunga2024} propose a RAG framework that is based on case-based reasoning. This work aims to highlight case-based reasoning as an adequate theoretical framework for legal question-answering (QA). Besides, the theoretical groundwork, the authors provide an implementation called CBR-RAG for the task of legal QA on the Australian Open Legal Corpus. 

Recent architectures tend to be more simplistic. \citet{Vijayakumaran2025} created a chatbot for real-time judicial insights that utilises FAISS \cite{Douze2024} along with MPNet \cite{Song2020} as the embedding model to create a chatbot that rapidly searches across legal data sources. \citet{AjayMukund2025} propose a dynamic legal RAG system that utilises a retrieval model based on the BM25 information retrieval algorithm \cite{Robertson1994} for the purpose of providing summarisation of Indian legal text. The authors have evaluated a variety of retrieval models, including more recently developed configurations and have resulted that BM25 is the most effective for their case. \citet{Schwarcz2025} conducted an empirical analysis on the use of RAG systems to enhance the quality of legal work of upper-level law students. Participants were divided into two groups, the first group utilised a system based on OpenAI's o1-preview reasoning model, while the other did not.

Over the past few years a number of assessment tools have been introduced examining different TAI assessment criteria, such as the Assessment List for Trustworthy AI (ALTAI) focusing on the implementation of the seven Trustworthy AI principles \cite{EC2020}, conformity assessments tools that ensure that AI systems meet regulatory standards like the European regulation for Artificial Intelligence, including capAI \cite{Floridi2022}, the MIT AI risk repository \cite{Slattery2024}, and impact assessment tools, including Human Rights Impact Assessment tools, such as Human Rights Impact Assessment Guidance and Toolbox of the Danish Institute for Human Rights \cite{DIHR2020HRIA}.

The mandate of AI Act has increased the urge for tools that enable its compliance. In turn, tools for legal AI assessment have emerged. We report some indicative ones. The impact assessment tool for public authorities \cite{Loi2021} is a two-stage checklist-based assessment that aims to achieve human-in-the-loop legal and ethical alignment. The AI Act Conformity Tool of Digital SME Alliance \cite{DSME2025} and the EU AI Act Compliance Checker \cite{AIACTCompl2024} offer both a web-based interactive questionnaire that guides the process of complying with the AI Act. The systems involved within these tools are mainly rule-based and are produced through a set of questions that the user fills in. The system proposed in this paper, differentiates by eliminating questions and requiring only minimal input, specifically the features of the AI system under assessment. Unlike rule-based approaches that are manual, error-prone, and difficult to maintain, our method relies on automated retrieval from the original document to produce the assessment.

\section{Development and Architecture}
This section presents the design philosophy, architecture, and core features of the proposed tool, developed within the context of the EU-funded DIGITAL-CSA Europe project called DeployAI. DeployAI aims to build and launch the AI-On-Demand Platform (AIoDP) to promote trustworthy, ethical, and transparent European AI solutions for use in industry, mainly for SMEs, and the public sector. AIoDP will support rapid prototyping and deployment of TAI applications across cloud, edge, and high-performance computing infrastructures.

Therefore, addressing the needs and requirements of the relevant key stakeholders (including  industrial stakeholders, researchers, policy-makers among others) is crucial, which aligns with the user-centric approach that has been adopted for the design of the tool. Feedback regarding their requirements has been received through frequent consultations. A key requirement that has emerged is the need for a self-assessment tool with the following features: a tool that enables compliance with the relevant EU AI regulations, that requires minimal input from the user, and thus to facilitate the ease of use. Additionally, to provide trusted results in a quick and effective assessment process, and lastly, to provide an initial guidance for the users transparently. As such, our proposed system architecture aims to address these requirements, while also developing a modular solution that enables the seamless software maintenance of all components. At the same time, it facilitates the technical robustness of the architecture and enables the autonomous deployment and continuous development of all its individual components. In turn, this approach allows for a scalable deployment to cloud services or locally hosted infrastructures. Each component is containerised and communicates with the rest via REST web APIs. Practically, this means that our architecture can be accelerated with upscaling of hardware and larger infrastructures. 

\begin{figure}[h]
    \centering
    \subfigure[System Architecture]{\includegraphics[width=0.48\textwidth]{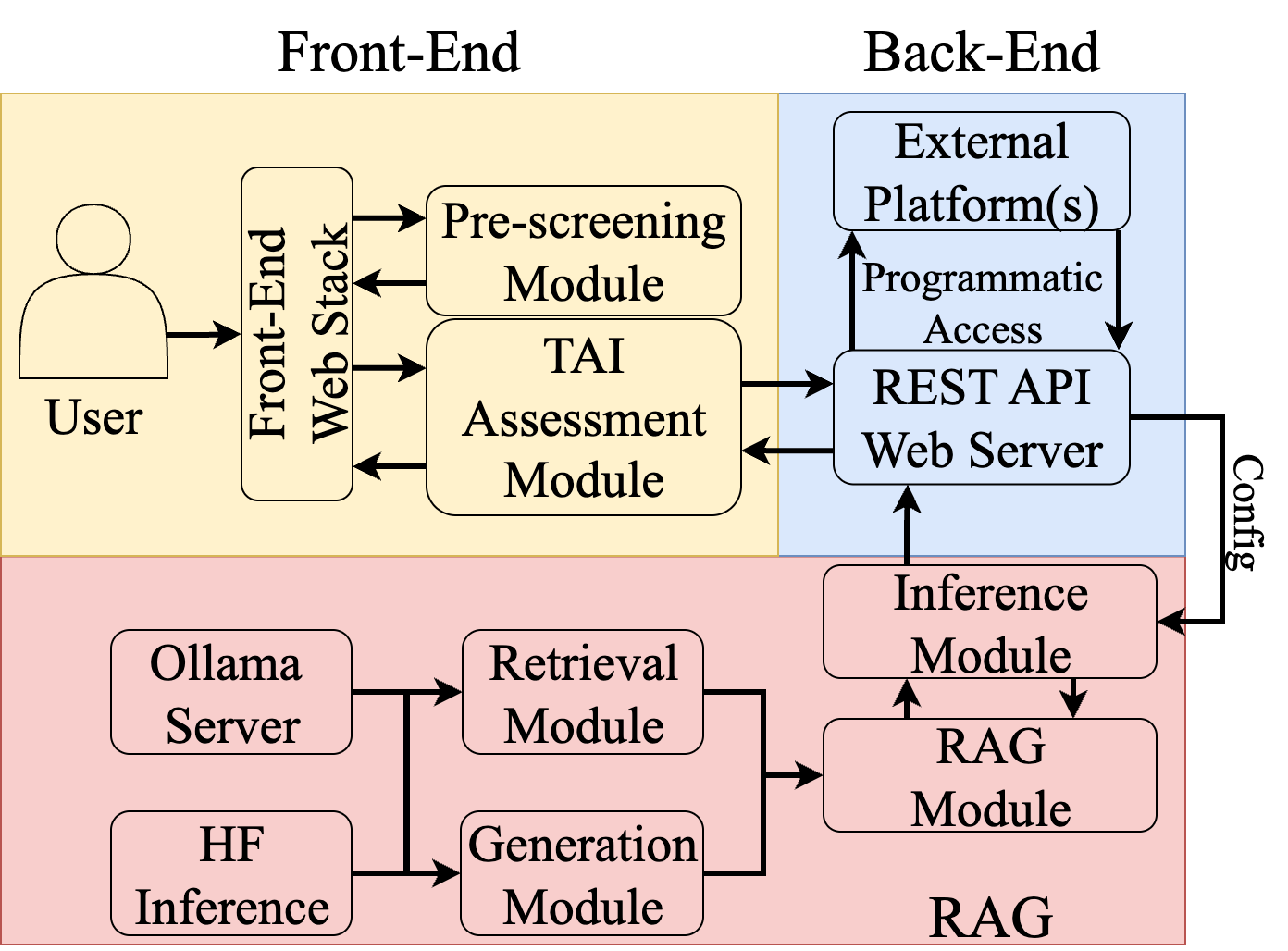}}
    \subfigure[TAI Assessment Front-End Screen]{\includegraphics[width=0.49\textwidth]{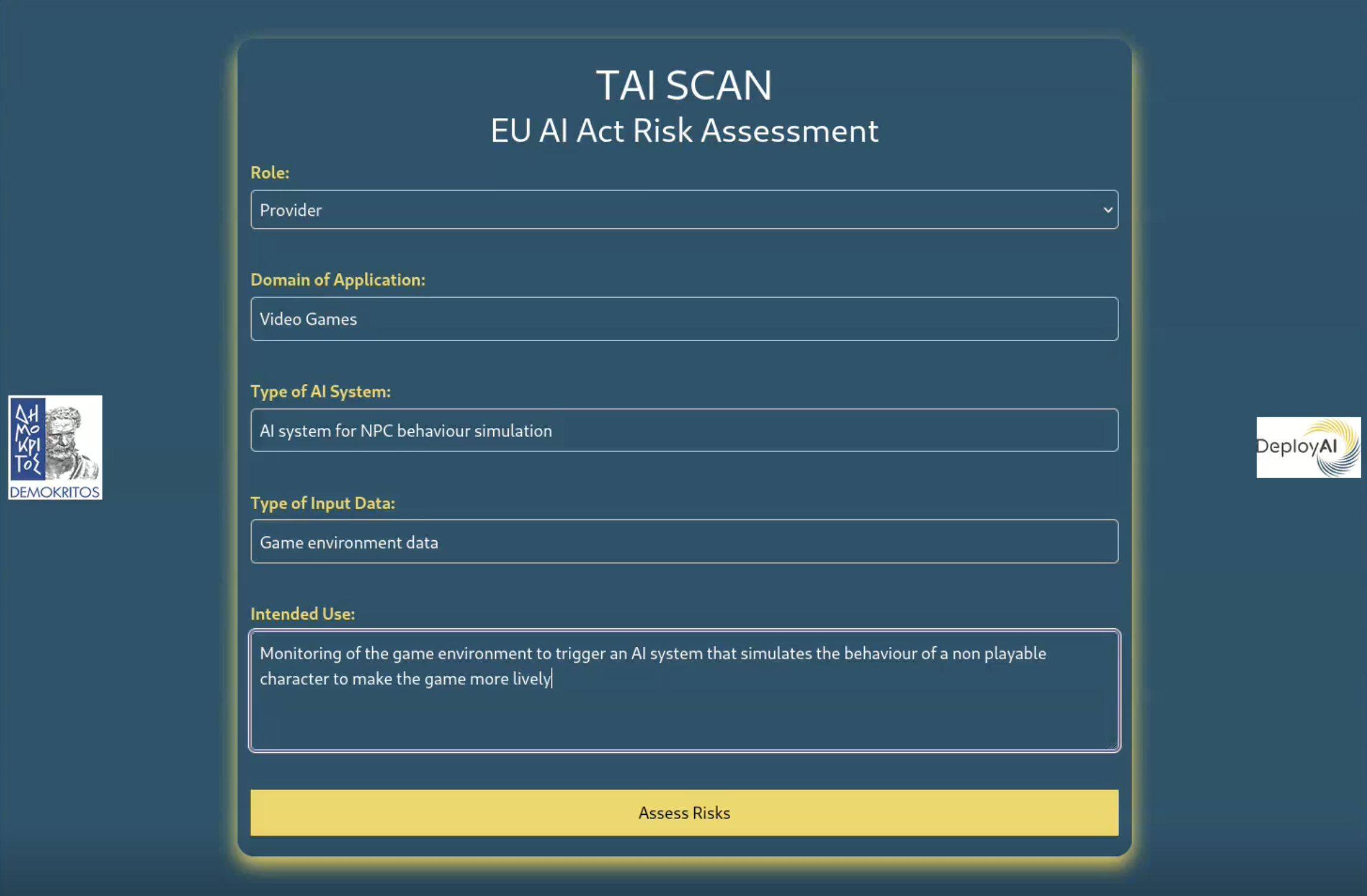}}
    \caption{System architecture diagram (a) illustrates the core components and the data flow. Top-left rectangle represents the modules that take place in front-end, while top-right rectangle represents back-end modules and their connection to external platforms. Lastly, bottom rectangle represents the RAG architecture. Subfigure (b) showcases the front-end for the TAI assessment trigger that initiates the process of assessment and communicates with the rest of the components.}
    \label{fig:architecture}
\end{figure}

Figure \ref{fig:architecture} portrays the architecture of the system, which consists of three distinct parts. First, the front-end, which is the interactive component. The user interacts with the front-end web stack that switches the flow between the pre-screening and the TAI assessment modules. The pre-screening module consists of a set of carefully curated questions that aim to notify and educate the user regarding prohibited or high-risk practices, which systems may be considered as AI systems under the EU AI law, and systems that may be classified as GPAI. Besides, it also acts as a safety measure against involvement of AI systems with non-acceptable properties. To proceed to the legal TAI assessment, the user has to provide an answer configuration that does not involve either prohibited or high-risk practices. Upon successful pre-screening, the screen switches to facilitate the input of the features of the AI system under assessment, with that signifying a TAI assessment trigger. A listener in the form of a REST API handles these requests. This listener can also act as an end-point to other external platforms, such as the AIoDP, and thus enabling programmatic access to the tool. To respond to the TAI assessment request, the REST API then initiates the inference module of the RAG component, which consists of the retrieval and generation modules. The retrieval module retrieves the most relevant document parts from the knowledge base and provides them to the generation module to output the assessment results. Both retrieval and generation modules can handle either the Ollama or HuggingFace backend.  Ollama allows for remote or local inference, while HuggingFace inference is currently only set for local inference.

The core features and functionality of the TAI Scan Tool consists of a two-stage assessment approach, namely (a) the pre-screening and  (b) the TAI assessment. Tables \ref{tab:preio]} and \ref{tab:taiassio} display the input and output of the core functionalities of the TAI Scan Tool. 

\begin{table}[t]
  \centering
  \caption{Input and output of the pre-screening feature.}
  \label{tab:preio]}
  \begin{tabular}{p{0.35\linewidth}p{0.55\linewidth}}
  \toprule
  \multicolumn{1}{c}{Input (Checkboxes)} & \multicolumn{1}{l}{Output} \\
  \midrule
  Does your technological system meet ALL of the following criteria? & 
    \multirow{6}{=}{Classification:
                      \vspace{-0.01\textwidth}
                      \begin{itemize}
                        \item AI System under the AI Act
                        \vspace{-1ex}
                        \item Not an AI system under the AI Act
                        \vspace{-1ex}
                      \end{itemize}
                      Risk Level:
                      \vspace{-0.01\textwidth}
                      \begin{itemize}
                        \item Prohibited AI system -- can not be deployed
                        \vspace{-1ex}
                        \item High-Risk -- strict requirements apply
                        \vspace{-1ex}
                        \item Not High-Risk
                      \end{itemize}
                      GPAI:
                      \vspace{-0.01\textwidth}
                      \begin{itemize}
                        \item Yes -- further assessment needed
                        \vspace{-1ex}
                        \item No -- not applicable
                        \vspace{-1ex}
                      \end{itemize}
    } \\
    Does your system conduct any of the following prohibited activities? & \\
    Is your AI system related to Union Harmonisation Legislation? & \\
    Is your AI system used in any of the following high-risk applications? & \\
    Does your AI system meet any exemption condition? & \\
    General Purpose AI (GPAI)? & \\
    \bottomrule
  \end{tabular}
\end{table}

\begin{table}[h]
  \caption{Input, input type, and output of the legal TAI assessment feature.}
  \label{tab:taiassio}
  \begin{tabular}{ccp{0.60\linewidth}}
    \toprule
    \multicolumn{1}{c}{Input} & \multicolumn{1}{c}{Input Type} & \multicolumn{1}{l}{Output} \\
    \midrule
    Role & Dropdown & \multirow{5}{=}{AI Act Risk Level: [Low-Risk, Medium-Risk, High-Risk, Prohibited] \\
                      Relevant References:
                      \vspace{-0.01\textwidth}
                      \begin{itemize}
                        \item Articles: [set of relevant articles]
                        \vspace{-1.5ex}
                        \item Recitals: [set of relevant recitals]
                        \vspace{-1.5ex}
                        \item Annexes: [set of relevant recitals]
                      \end{itemize}
    } \\
    Domain of application & Free text & \\
    Type of AI system & Free text & \\
    Type of input data & Free text & \\
    Intended use & Free text &                         \vspace{1.8ex}\\
  \bottomrule
\end{tabular}
\end{table}

Depending on the selected checkboxes, the pre-screening function provides output and insight on whether the system is an AI system under the AI Act, whether is prohibited, high-risk or not high-risk, and whether it is a GPAI system. The main focus is to safeguard against non-acceptable practices. However, it can also act as a rapid way to receive feedback on the AI Act classification of an AI system. While most developers have an inherent perception of the risk level of their system, this does not always align with the AI Act. Due to its functionality operating on the front-end, users have greater flexibility on how they interact with the system without impacting their system's assessment. To proceed with further assessment the user has to select a configuration that (a) classifies his system as an AI system under the AI Act and (b) does not include high-risk or prohibited practices. 

TAI assessment provides a more comprehensive assessment of the AI system. Therefore, it requires a particular set of features such as its role (provider/deployer) according to the AI Act, the domain of application, the type of AI system, the type of input data and the intended use. Most of the input is free text. Given the above features, the RAG system provides the risk level of the system and the relevant references from the document to aid on compliance with the respective framework. Currently, only the legal compliance is provisioned with a strong focus on the AI Act. Next version of the tool will feature more documents and TAI components. 

\section{Evaluation and Results}
\label{sec:eval}
This section presents the qualitative evaluation of the TAI Scan Tool and the interpretation of findings. As previously mentioned, the current version of the tool includes the legal compliance on AI Act. The evaluation includes a set of scenarios from different risk levels according to the AI Act. These scenarios enable the exploration and limits of the reasoning of the system. At the same time, it acts as a seamless manner to grasp the contextual factors and the document parts that influence the outcomes of the generation module. Besides, this process is complementary, and it lays the groundwork for a more robust quantitative evaluation. 

\begin{table*}[h]
  \caption{Qualitative Evaluation With Scenarios along with the description and output from the TAI Scan Tool.}
  \label{tab:qual-eval}
  \begin{tabular}{p{0.13\linewidth}p{0.28\linewidth}cl}
    \toprule
    Risk-Level (Role) & Scenario Description & Predicted Level & Relevant Articles \\
    \midrule
    Prohibited (Provider) & Real-time remote biometric identification & Prohibited & [14, 13, 26, 12, 49, 16, 9, 6, 5, 27] \\
    High-Risk (Provider) & AI-driven digital infrastructure management system & High-Risk & [13, 14, 9, 12, 27, 15, 17, 8, 42] \\
    High-Risk (Deployer) & AI-driven digital infrastructure management system & High-Risk & [13, 14, 9, 12, 27, 16, 26, 15, 8, 49]  \\
    Low-Risk (Provider) & Video game NPC behaviour & Low-Risk & [13, 14, 9, 15, 16, 8, 6, 42, 12, 10] \\
  \bottomrule
\end{tabular}
\end{table*}

Table \ref{tab:qual-eval} includes the scenarios tested for our evaluation. Our set of scenarios consists of one prohibited, two high-risk and a low-risk level scenario. To create these scenarios, we advised the relevant articles from the AI Act. Article 5 includes the setting of prohibited AI practices, while Article 6 includes the classification rules for high-risk AI systems. In addition, Annex III supports Article 6 by defining additional configuration settings of high-risk systems. The success criteria for our evaluation are mainly for the predicted risk level to be aligned with the ground truth. Furthermore, we aim to draw insights regarding the contextual factors that impact the decisions of our system. To explore this, we manually navigate and check the title and content of each article and its relevance to the predicted level and the scenario description.

Following is the general configuration of the RAG component. First, we turn the AI Act articles into embeddings using the Jina AI BERT-based embedding model\footnote{https://huggingface.co/jinaai/jina-embeddings-v3}. We save the vector database locally to compare with the query. To match the query with the relevant articles we transform the query into the embedding space using the same model. We use a medium-sized model from Mistral, called mistral-small3.2\footnote{https://ollama.com/library/mistral-small3.2} to rewrite the query, based on an empirically tested rubric, to achieve optimal search relevance and query expansion.  We employ Spotify's Annoy algorithm\footnote{https://github.com/spotify/annoy} which is a GPU enabled approximate nearest neighbour implementation. Finally, we prompt mistral-small3.2 to generate a response to the query, given the relevant articles as context. The prompt is customised for the purposes of standardised output. According to \citet{Gao2023}, this is classified as an advanced RAG configuration, with pre-retrieval techniques. Recent advancements both in the domain of feature extraction and natural language generation have been increasing the capacity of models and alleviating frequent challenges. In turn this enables the design of more simplistic configurations.

We conducted our evaluation by utilising the back-end component of our system to automatically and programmatically inference the RAG component. To automate the evaluation process we built several configuration files that contained relevant input and output directories, hyperparameters, seeds and specialised queries used for the retrieval, operation, and inference of RAG. This functionality not only allowed us automation but also demonstrates the effectiveness of the tool as a component that can be integrated in larger TAI governance frameworks. The low-risk scenario is the only scenario that could be tested using the front-end, as pre-screening would not allow the user to proceed to further TAI legal scan.

\subsection{General Trends}
Our system correctly predicts the risk level of all tested scenarios. We observe that all four sets of relevant articles include the following articles: 9, 12, 13, 14. Article 9 title ``Risk Management System'' mandates the establishment and maintenance of a lifecycle-wide risk management system for providers of high-risk AI systems. Article 12 titled ``Record-Keeping'' reports that an automatic recording of events should take place on high-risk AI systems to ensure traceability appropriate to their intended use. Article 13 titled ``Transparency and Provision of Information to Deployers'' states that high-risk AI systems should cover the transparency aspect of their systems in order for the end-users to understand them and use them appropriately. Lastly, Article 14 titled ``Human Oversight'' highlights that the design of the high-risk AI systems should be intuitive in a way that would allow the effective overseeing from humans. 

By analysing the titles and content of each article, we observe that these articles act as general guidelines on high-risk AI systems and act as horizontal obligations. However, we observe two additional trends. First, the retrieval part focuses on articles related to high-risk AI systems. We hypothesize that this happens due to the setting of high-risk AI systems, as well as, the terms related to this risk-level being statistically more frequent in the data distribution. Besides, high-risk systems require careful consideration and monitoring to be deployed and used safely, therefore it is expected to require clear definition and representation in the AI Act. Therefore, comparison to the setting of high-risk AI systems is an approach to creation of a decision boundary. Furthermore, we observe an overlap between the roles of providers and deployers. This is evident by retrieval of Article 16 during the deployer scenario. This is a side-effect of Article 50 that is titled ``Transparency Obligations for Providers and Deployers of Certain AI Systems'' that interconnects these two different terms together. Lastly, we observe that our system displays three distinct groups of relevant articles which can be title as following: horizontal obligations, other related obligations and classification resources. 

\subsection{Interpretation of Findings}
In this section we analyse further our findings. Table \ref{tab:arts} displays all the relevant articles found in the output of our system along with their titles to aid the comprehension of our interpretation.

\begin{table}[t]
  \caption{AI Act articles found in our qualitative evaluation along with their titles.}
  \label{tab:arts}
  \begin{tabular}{cp{0.35\linewidth}cp{0.40\linewidth}}
    \toprule
    Article & Title & Article & Title\\
    \midrule
    5 & Prohibited AI Practices & 15 & Accuracy, Robustness and Cybersecurity \\
    6 & Classification Rules for High-Risk AI Systems & 16 & Obligations of Providers of High-Risk AI Systems \\
    8 & Compliance with the Requirements & 17 & Quality Management System \\
    9 & Risk Management System & 26 & Obligations of Deployers of High-Risk AI Systems \\
    10 & Data and Data Governance & 27 & Fundamental Rights Impact Assessment for High-Risk AI Systems \\
    12 & Record-Keeping & 42 & Presumption of Conformity with Certain Requirements \\
    13 & Transparency and Provision of Information to Deployers & 49 & Registration \\
    14 & Human Oversight & & \\
  \bottomrule
\end{tabular}
\end{table}

As mentioned previously all sets of relevant articles for all scenarios follow the three groups mentioned earlier. For the prohibited scenario, we observe that articles 5 and 6 are utilised by the system to infer/classify the scenario into one of the risk-levels as they contain the setting for prohibited and high-risk scenarios. Articles 9, 12, 13, and 14 as mentioned in the previous section can be considered as the horizontal obligations group. Lastly, Articles 16 and 26 portray the obligations for providers and deployers of high-risk systems. Due to remote biometric use on public spaces this goes against fundamental rights, and thus why article 27 is relevant. The above premise indicates the need for registration of indicated by article 49.

Moving on to the high-risk scenarios, we observe a similar trend, however the output of the system diversifies for the different roles of provider and deployer. Besides, the horizontal obligation articles, article 15 that involves the setting for cybersecurity acting as the classification article due to involving digital infrastructures. The scenario specific obligations suggest the mandate of a quality management system, compliance with high-risk specific requirements, securing fundamental rights due to operating on critical infrastructures, and conformity to use with test and train data focusing on their intended use. The deployer scenario, includes reference to article 26 that describe the obligations of deployers and need for registration.

Finally, the low-risk scenario outputs the horizontal obligations, along with articles 6 and 15 for classification and articles 8, 10, 42 and 16 as scenario specific obligations. Due to the scenario being low-risk we hypothesize that all the articles are utilised as resources for correct classification. As mentioned in the previous section, the system portrays the following behaviour, it tries to understand the setting and create a comparative profile of high-risk systems, due to the reference frequency of high-risk systems in the AI Act. Then it compares the high-risk setting to that of the scenario provided and produces its risk-level by also taking into account all the relevant obligations.

\section{Future Work and Conclusions}
In this paper we have introduced the TAI Scan Tool, a multi-component system that rapidly provides a TAI assessment for AI systems. The current version provides an assessment for the legal perspective of TAI with a focus on the AI Act compliance. Our tool differentiates from the similar tools for TAI assessment and particular AI Act compliance tools by requiring minimal input for the process of legal TAI assessment. Furthermore, we deploy a data-driven approach based on the RAG framework for document grounded reasoning, that reduces manual effort and is easily extensible with other documents and TAI components.

Qualitative evaluation of our system displays correct prediction of risk-level and high contextual relevance of the retrieved articles. Our system portrays two general trends. First, the proposed relevant articles of our system can be categorised into three distinct groups: (i) horizontal obligations, (ii) classification rules, (iii) scenario specific obligations. These three groups are utilised to make a decision regarding the risk-level of the AI system. We also observe that the setting of high-risk systems, their related terms and definitions are very frequent along the span of the document. This is expected as high-risk systems require careful deployment, monitoring, and recording in order to be operating safely. This is also reflected in the data distribution and the embeddings created by the retrieval component of RAG. 

Therefore, our system constitutes an effective and rapid TAI assessment that can take place during all phases of the AI development lifecycle. The above process could take place on voluntary basis as an internal assessment mechanism of any organisation developing AI systems. It could be of particular interest to SMEs or startups that are not able to allocate budget for legal consultations or do not have legal departments. Due to our modular approach it can be integrated into larger platforms or internal governance frameworks with the aim to carefully monitor and notify the platform administrators for any prohibited or high-risk systems about to be onboarded. 

This paper has laid the groundwork for the TAI Scan Tool, future work will focus on expanding the current version with the following directions. First, introducing more documents in the knowledge base of legal TAI assessment. Furthermore, we will focus on expanding with the rest of the TAI components, i.e., ethical and robust. Although initial qualitative results are prominent, our next steps will be on deploying quantitative evaluation to test the effectiveness of our approach. The current naive RAG requires a low capacity of computational resources thanks to the selection of models. However, to address any challenges that will yield during quantitative evaluation elevating other well-received techniques, such as reranking of context, among others, will be prioritised.
 
\paragraph{Acknowledgements} 
This work has been funded by the Digital Europe Programme (DIGITAL) under grant agreement No. 101146490 - DIGITAL-2022-CLOUD-AI-B-03.
\bibliography{main}

\end{document}